\documentclass{article}
\usepackage{spconf,amsmath,graphicx}
\usepackage[ruled,vlined]{algorithm2e}
\usepackage{booktabs}
\usepackage[flushleft]{threeparttable}
\SetKwComment{Comment}{$\triangleright$\ }{}
\newcommand{\olsi}[1]{\,\overline{\!{#1}}}
\newtheorem{definition}{Definition}[section]

\title{Federated marginal personalization for ASR rescoring}
\name{Zhe Liu, Fuchun Peng}
\address{Facebook AI, Menlo Park, CA, USA}

\begin{document}
%
\maketitle
\begin{abstract}
We introduce federated marginal personalization (FMP), a novel method for continuously updating personalized neural network language models (NNLMs) on private devices using federated learning (FL). Instead of fine-tuning the parameters of NNLMs on personal data, FMP regularly estimates global and personalized marginal distributions of words, and adjusts the probabilities from NNLMs by an adaptation factor that is specific to each word. Our presented approach can overcome the limitations of federated fine-tuning and efficiently learn personalized NNLMs on devices. We study the application of FMP on second-pass ASR rescoring tasks. Experiments on two speech evaluation datasets show modest word error rate (WER) reductions. We also demonstrate that FMP could offer reasonable privacy with only a negligible cost in speech recognition accuracy.
\end{abstract}
\section{Introduction}
\label{sec:intro}

In recent years, there has been a rise in the popularity of a distributed learning technique called \emph{federated learning} (FL) \cite{konevcny2016federated, konevcny2016federated2, mcmahan2017communication, geyer2017differentially}. It protects the privacy of data by training a shared machine learning model in a decentralized manner on users' local devices, so that raw data never leaves physical devices. Each client model takes a global model from the central server for parameters initialization, and trains its own private local model using personal data. FL has been applied in many fields including recommendation \cite{chen2018federated}, keyboard suggestion \cite{arnold2016suggesting}, keyword spotting \cite{leroy2019federated}, phenotyping \cite{kim2017federated} and health care \cite{xu2019federated}.

Among these applications, \emph{language modeling} is one of the most common tasks and serves as an important module in automatic speech recognition (ASR) \cite{goodman2001progress}. In particular, neural network language models (NNLMs) typically outperform traditional $n$-gram language models in better keeping track of long range dependency \cite{mikolov2010recurrent,chen2015improving}, and are widely used in the second-pass decoding via $N$-best or lattice rescoring \cite{xu2018pruned}.

A common issue arising after deploying an ASR model on user device is the discrepancy between training data and actual  data received on local devices. In the case of language modeling, the language and style of real users' utterances can be very different from those of generic training corpus. The most general method to address this challenge is integrating a separate \emph{personalized} language model trained on device in the FL framework. Particularly, the method of \emph{federated fine-tuning} on private data has been explored in recent literature \cite{brendan2018learning, popov2018distributed, ji2019learning, chen2019federated}, where we start with a general language model downloaded from the central server and have it continuously updated on devices using distributed parameter fine-tuning.

However, there exists several limitations confronting this federated fine-tuning method: (1) If each user only generates a very small number of utterances, such data by itself cannot be used for updating the general language model; (2) Fine-tuning a large pre-trained general language model on personal data tends to suffer from overfitting, or \emph{catastrophic forgetting} \cite{goodfellow2013empirical}; (3) Training neural models on user devices typically has resource constraints and could be computationally expensive.

In this paper, we introduce \emph{federated marginal personalization} (FMP), a novel approach for continuously updating personalized NNLMs on private devices using FL. Instead of fine-tuning the parameters of NNLMs on user personal data from first-pass ASR decoded words, FMP regularly estimates and updates global (server side) and personalized (client side) unigram distributions, and multiplies the probabilities from on-device NNLMs by a factor specific to each word. Then the resulting adapted language models are utilized in the second-pass ASR rescoring. Our proposed method can overcome the limitations of federated fine-tuning framework and efficiently learn personalized NNLMs. We also demonstrate that FMP satisfies utterance-level \emph{differential privacy} (DP) \cite{dwork2006calibrating, dwork2014algorithmic} with only a negligible cost in speech recognition accuracy.

The idea of leveraging word frequencies to bias language model probabilities was originally presented in \cite{kneser1997language}, further studied in \cite{singh2017approximated}, and authors in \cite{li2018recurrent} adopt such fast marginal adaptation framework to adapt recurrent neural network language models. In our work, we continue this line of research and take advantage of FL to interpolate global and personal marginals of word distributions for on-device NNLM personalization. To the best of our knowledge, our paper is the first one that leverages FL to explore fast marginal language model personalization with its application in ASR rescoring.

The rest of the paper is organized as follows. In Section~\ref{sec:method}, we introduce the FMP approach for ASR rescoring tasks. We evaluate the proposed method in Section~\ref{sec:expt} and demonstrate privacy analysis in  Section~\ref{sec:privacy}. We conclude in Section~\ref{sec:summary}.

\section{Methods}
\label{sec:method}
In this section, we describe the FMP approach on second-pass ASR rescoring tasks. To start with, we train a general NNLM $p_{\text{nnlm}}(w|h)$ using background corpus on the server side, and deploy this model to each physical device for second-pass $N$-best rescoring. Here, $w$ stands for any word and $h$ represents the context history. Let $u(w)$ be the discrete unigram distribution estimated from the background corpus. It is delivered to each local device as well along with the initial deployment of ASR model and second-pass NNLM rescorer.

Our approach can be outlined in Algorithm~\ref{algo}, with details provided in Sections~\ref{sec:client}, \ref{sec:per} and \ref{sec:server}.

\begin{algorithm}
\label{algo}
\SetAlgoLined
 hyper-parameters $\lambda, \alpha, \beta, \sigma$\;
 initialize $p_{\text{nnlm}}(w|h), u(w)$\;
 \For{\emph{each FL round} $t=0,1,2,\ldots$}{
  \For{\emph{each client} $i=1,2,\ldots, n$}{
    \eIf{$t = 0$}{
     $p_{i, \text{nnlm}}^{t}(w|h)\leftarrow p_{\text{nnlm}}(w|h)$\;
    }{
     $p_{i, \text{nnlm}}^{t}(w|h)\leftarrow$ according to Eq.~(\ref{eq:fmp});
    }
    conduct ASR rescoring using $p_{i, \text{nnlm}}^{t}(w|h)$\;
    $q_i^t(w)\leftarrow$ according to Eq.~(\ref{eq:qi})\;
    $c_i^t\leftarrow$ according to Eq.~(\ref{eq:c})\;
  }
  $\olsi{q}^{t}(w)\leftarrow$ according to Eq.~(\ref{eq:fa})\;
 }
 \caption{FMP approach for ASR rescoring.}
\end{algorithm}

\subsection{Client-side model update}
\label{sec:client}
Each client $i$ receives the global unigram distribution $\olsi{q}^{t}(w)$ from server by the end of round $t$ and performs the following update on client-side NNLM in round $t+1$
\begin{align}
\label{eq:fmp}
    p_{i, \text{nnlm}}^{t+1}(w|h)= \frac{1}{Z_i^t(h)}\cdot\left(\frac{g_i^t(w)}{u(w)}\right)^\lambda \cdot p_{\text{nnlm}}(w|h),
\end{align}
where
\begin{align}
    g_i^t(w):= (1-\alpha-\beta) \cdot u(w) + \alpha\cdot\olsi{q}^{t}(w)+\beta\cdot q_i^t(w).
\end{align}
Here, $q_i^t(w)$ is the personalized unigram distribution for user $i$ estimated from ASR decoded text in an unsupervised manner, hyper-parameter $\lambda\geq 0$ controls the scaling factor of marginal adaptation, and $Z_i^t(h)$ is a normalization constant. We defer the estimation method of $q_i^t(w)$ to Section~\ref{sec:per}.

Notice that the numerator of the scaling factor, $g_i^t(w)$, is a linear interpolation of background unigram estimates $u(w)$, global unigram estimates $\olsi{q}^{t}(w)$, and personalized unigram estimates $q_i^t(w)$, with interpolation weights of $1-\alpha-\beta\geq 0$, $\alpha\geq 0$ and $\beta\geq 0$, respectively. Here, our intuition is that the updated NNLM should respect the general words, \emph{in-domain} or globally trending words, as well as personalized words that are particularly uttered by user $i$. For example, in voice search applications, once initial models are shipped to local devices, we use global unigram distribution to adapt live traffic, and personalized unigram distribution to account for user-level language and style. Also, we still require general background unigram distribution in the interpolation to prevent NNLMs from overfitting on the decoded text received on the devices.

For efficiency purposes in ASR rescoring tasks, we use an unnormalized version of Equation~(\ref{eq:fmp}) to adjust the NNLM output word probabilities during second-pass rescoring.

\subsection{Estimation of personalized unigram distribution}
\label{sec:per}
After a word is spoken, there is more chances that it is spoken again by the same user. For each user $i$, personalized unigram distribution $q_i^t(w)$ exploits the word distribution of historical context up to FL round $t$. In particular, we maintain a running ``cache'' that keeps track of the word counts from historical ASR decoded text, and estimate the corresponding unigram distribution by counts normalization and smoothing.

For unigram counting, we can leverage the decoded words from all the $N$-best hypotheses obtained from first-pass ASR decoding, with the use of \emph{Gaussian kernel} weighting
\begin{align}
    K(s) = \exp\left(-(\text{rank}(s) - 1)^2/(2\sigma^2)\right).
\end{align}
Here $s$ represents an ASR decoded hypothesis, $\text{rank}(s)$ stands for the rank of $s$ among the $N$-best list, and the bandwidth hyper-parameter $\sigma>0$ controls the weighting scale. Notice that as $\sigma$ approaches infinity, we approximately have uniform weights over $N$-best list; when $\sigma$ is close to zero, we put zero weights on all hypotheses except the 1-best hypothesis.

Personalized unigram distribution can be estimated by
\begin{align}
\label{eq:qi}
    q_i^t(w) &= \text{Smoothing}\bigg(\sum_{s\in S_i^t} K(s)\cdot c_s(w)\big/c_i^t\bigg), \\
    c_i^t:&=\sum_{w}\sum_{s\in S_i^t} K(s)\cdot c_s(w), \label{eq:c}
\end{align}
where $S_i^t$ contains all the hypotheses generated from user $i$ by the end of round $t$, and $c_s(w)$ computes the count of word $w$ among the hypothesis $s$. It is worth noting that in practice the personalized unigram distribution $q_i^t(w)$ can be estimated continuously to allow finer personalization, for example, it can be refreshed every time that the ASR model transcribes a new utterance on device, instead of being updated only once per each FL round.

In our application of second-pass ASR rescoring, NNLMs adapt the text data that is labeled by first-pass ASR decoding in an unsupervised manner. An alternative approach is to take advantage of the \emph{soft labels} predicted by ASR models \cite{shin2016generative} and estimate unigram distributions accordingly. This method is beyond the scope of this study.

\subsection{Server-side model update}
\label{sec:server}
After on-device estimation for round $t$, these locally updated unigram distributions $q_i^t(w)$ are sent to the central server for global aggregation. We adopt the following global update rule of \emph{federated averaging} \cite{mcmahan2017communication}
\begin{align}
\label{eq:fa}
    \olsi{q}^{t}(w)=\frac{\sum_{i=1}^n c_i^t q_i^t(w)}{\sum_{i=1}^n c_i^t},
\end{align}
where $q_i^t(w)$ is the estimated unigram distribution for user $i$ and round $t$, and $c_i^t$ is the corresponding sum of word pseudo-counts which serves as the weight for averaging. Notice that after updating, the global unigram distribution $\olsi{q}^{t}(w)$ will be sent back to local devices and utilized in round $t+1$.

In practical implementation, instead of sending each local personalized unigram distribution to the server, user devices can send distribution deltas \cite{bui2019federated}, i.e., the difference between current personalized distribution and the global distribution before updating. Moreover, in real-world applications where there is a large number of user devices, we typically sample only a subset of users before performing federated averaging.


\section{Experiments}
\label{sec:expt}
\subsection{Datasets}
In our experiments, the first-pass ASR model is trained using the in-house video ASR datasets (14K hours), which are sampled from public social media videos and de-identified before transcription; both transcribers and researchers do not have access to any user-identifiable information (UII). For second-pass rescoring, the general background text that we use to train NNLM is a corpus of public Facebook posts and comments, which contains around 30M English sentences.

We evaluate the proposed method on two speech datasets. The first is a \emph{curated} set of carefully select very clean videos. Each video is segmented into multiple chucks of utterances. The second dataset is the \emph{Augmented Multi-Party Interaction} (AMI) Meeting data \cite{carletta2005ami}. It includes scenario meetings (with roles assigned for participants) and non-scenario meetings (where participants were free to choose topics). For scenario meetings, each session is divided into 4 one-hour meetings. Each meeting has 4 participants. The sizes of these datasets are summarized in Table~\ref{tab:data}.

\begin{table}[ht]
  \caption{Summary of speech evaluation datasets.}
  \centering
  \resizebox{\columnwidth}{!}{%
  \begin{tabular}{l|r|r}
    \toprule
    & \multicolumn{2}{|c}{\bf{Evaluation Dataset}} \\
    \cmidrule(r){2-3}
    \emph{Feature}& \emph{Curated Video} & \emph{AMI Meeting}  \\
    \midrule
    Num.~of videos/meetings & 78 & 63 \\
    Num.~of utterances & 1,015 & 12,643 \\
    Num.~of words & 74,248 & 89,666 \\
    \bottomrule
  \end{tabular}
  }
  \label{tab:data}
\end{table}

\subsection{Setups}
For the first-pass ASR model, we use connectionist temporal classification (CTC) \cite{graves2006connectionist} criterion to learn an encoder-only model and is further composed with a 5-gram language model in a standard weighted finite-state transducers (WFST) framework. Here we adopt a latency-control bi-directional LSTM (LC-BLSTM) encoder with 6 layers of 1000 hidden units. For second-pass rescoring, we utilize a Transformer \cite{vaswani2017attention} based language model with word embeddings dimension 256, feed-forward network (FFN) dimension of 1024, 3 decoder blocks, 4 attention heads, and dropout of 0.10.

In our experiments, the baseline method is the first-pass ASR decoding with second-pass $N$-best rescoring using the general background NNLM. For each utterance, we generate 100-best hypotheses for \emph{Curated Video} and 20-best hypotheses for \emph{AMI Meeting} datasets. For each hypothesis, its NNLM score is linearly combined with the score from  first-pass 5-gram language model using interpolation weight 0.50.

To simulate the server-client environments for evaluating FL based approaches, we treat each video in \emph{Curated Video} data (or each meeting in \emph{AMI Meeting} data) as a client, and any utterances that belong to the same video (or meeting) are thus considered as being received on devices and transcribed by client-side first-pass ASR model with second-pass NNLM rescoring. Utterances from the same video (or meeting) are ranked based on the starting timestamp of recording. For the proposed FMP method with total number of FL rounds being $T$ (varied in our experiments), we evenly partition utterances from the same video (or meeting) into $T+1$ groups, and assume all utterances from group $t$ are received and processed in round $t-1$ of FL, where $t=1, \ldots, T+1$.

\subsection{Evaluation results}
\label{sec:res}
We evaluate the proposed FMP method by ASR rescoring task on \emph{Curated Video} and \emph{AMI Meeting} datasets. We set hyper-parameters $\alpha=0.5, \beta=0.25$, and $\sigma=5.0$; $\lambda$ is tuned on a small validation set and kept as the same across all our experiments. We measure the impact of different choices of hyper-parameters in Section~\ref{sec:compare}. Table~\ref{tab:eval} presents the word error rate (WER) results of FMP with various numbers of FL rounds $T$ comparing to the baseline approach. We can see that FMP improves WERs consistently on both datasets (relatively 2.4\% gain on \emph{Curated Video} and 4.8\% gain on \emph{AMI Meeting} datasets). In particular, the improvement becomes slightly larger as $T$ increases, which is expected since we can better leverage global marginals by more frequent updates.

\begin{table}[ht]
  \caption{WERs from the baseline and FMP methods.}
  \centering
  \resizebox{\columnwidth}{!}{%
  \begin{threeparttable}
  \begin{tabular}{l|l|l}
    \toprule
    & \multicolumn{2}{|c}{\bf{Evaluation Dataset}} \\
    \cmidrule(r){2-3}
    \emph{Method}& \emph{Curated Video} & \emph{AMI Meeting}  \\
    \midrule
    Baseline & 7.85 & 32.83 \\
    \midrule
    FMP w/ FL rounds $T=1$ & 7.67\;(-2.3\%) & 31.44\;(-4.2\%) \\
    FMP w/ FL rounds $T=2$ & \textbf{7.66}\;(-2.4\%) & 31.34\;(-4.5\%) \\
    FMP w/ FL rounds $T=5$ & \multicolumn{1}{c|}{-\tnote{*}} & 31.27\;(-4.8\%) \\
    FMP w/ FL rounds $T=10$ & \multicolumn{1}{c|}{-} & \textbf{31.24}\;(-4.8\%) \\
    \bottomrule
  \end{tabular}
  \begin{tablenotes}
    \footnotesize
    \item[*] Not evaluated due to small number of utterances in each FL round.
  \end{tablenotes}
  \end{threeparttable}
  }
  \label{tab:eval}
\end{table}

\subsection{Comparison of different hyper-parameters}
\label{sec:compare}
Table~\ref{tab:compare} shows the WERs comparison results among various hyper-parameters of the proposed FMP method ($T=2$ for \emph{Curated Video} data and $T=10$ for \emph{AMI Meeting} data). For the interpolation weights $\alpha$ and $\beta$ in marginal adaptation, we can see that both of them play their crucial roles and perform better than only using either of them. Regarding the Gaussian kernel bandwidth $\sigma$, we can see that $\sigma=1$ achieves the best results on both datasets, better than only using the 1-best hypothesis ($\sigma=0.1$) or all $N$-best hypotheses with equal weights ($\sigma=100$) to estimate personalized distributions.

\begin{table}[ht]
  \caption{WERs comparison among various hyper-parameters (interpolation weights $\alpha$, $\beta$; bandwidth $\sigma$) of FMP method.}
  \centering
  \resizebox{\columnwidth}{!}{%
  \begin{tabular}{l|l|c|c}
    \toprule
    \multicolumn{2}{c}{} & \multicolumn{2}{|c}{\bf{Evaluation Dataset}} \\
    \cmidrule(r){3-4}
    \multicolumn{2}{l|}{\emph{Hyper-Parameters of FMP}} &\emph{Curated Video} & \emph{AMI Meeting}  \\
    \midrule
    &$\sigma=5$ & 7.66 & 31.24 \\
    \cmidrule(r){2-4}
    &$\sigma=0.1$ & 7.67 & 31.02 \\
    $\alpha=0.5$, $\beta=0.25$ &$\sigma=1$ & \textbf{7.65} & \textbf{30.96} \\
    &$\sigma=10$ & 7.66 & 31.38 \\
    &$\sigma=100$ & 7.67 & 31.42 \\
    \midrule
    $\alpha=0.0$, $\beta=0.75$ &$\sigma=5$ & 7.70 & 31.56 \\
    $\alpha=0.75$, $\beta=0.0$ &$\sigma=5$ & 7.72 & 31.31 \\    
    \bottomrule
  \end{tabular}
  }
  \label{tab:compare}
\end{table}

\section{Privacy Analysis}
\label{sec:privacy}
A differentially private mechanism enables the public release of model parameters with a strong privacy protection \cite{dwork2006calibrating, dwork2014algorithmic}.
\begin{definition}[DP]
\label{def:dp}
A randomized mechanism $\mathcal{M}$ with a domain $\mathcal{D}$ and range $S$ satisfies $(\epsilon,\delta)$-DP if for any two adjacent datasets $d,d'\in\mathcal{D}$ and for any subset $S\subseteq\mathcal{S}$, it holds that
\begin{align}
    P(\mathcal{M}(d)\in S)\leq e^\epsilon P(\mathcal{M}(d')\in S) + \delta.
\end{align}
\end{definition}
Here $d$ and $d'$ are defined to be \emph{adjacent} if $d'$ can be formed by adding or removing a single training example from $d$.

In order to achieve DP, some randomness must be introduced to the algorithm. We use the $Laplace$ mechanism \cite{dwork2008differential} which adds a Laplace noise to the output of a query function
\begin{align}
\label{eq:privacy}
    \olsi{q}_{\epsilon,\text{DP}}^{t}(w)&=\frac{\sum_{i=1}^n c_i^t q_i^t(w)+r_\epsilon^t(w)}{\sum_{i=1}^n c_i^t+\sum_w r_\epsilon^t(w)}, \\
    r_\epsilon^t(w)&\stackrel{\text{i.i.d.}}{\sim}\text{Laplace}(1/\epsilon),
\end{align}
where $\text{Laplace}(b)$ stands for a Laplace distribution with mean 0 and variance $2b^2$, and parameter $\epsilon$ controls the strength of privacy protection. Intuitively, a larger $\epsilon$ leads to stronger privacy protection, but can degrade the model accuracy. 

Note that count has \emph{sensitivity} 1, i.e. maximum difference in the query function output from adjacent datasets. For simplicity, we assume that any sensitive word is uttered at most once in any utterance. It is straightforward to show (see \cite{dwork2008differential}) that randomized FMP with the server-side update provided in Equation~(\ref{eq:privacy}) satisfies $(\epsilon,0)$-DP at utterance level.

It is worth noting that the definition of \emph{adjacent datasets} in Definition~\ref{def:dp} depends on the application. Most prior work on DP deals with example level (or utterance level in our case). For ASR tasks, a better definition is user-level adjacency for protecting whole user histories in the training set \cite{brendan2018learning}, since a sensitive word may be uttered several times by an individual user. In such case, we need counting capping to give upper bounds of sensitivity. We leave this for future work.

We evaluate our approach using $\epsilon$ values from 0.1 to 2.0. Here, we use the 1-best hypothesis for personalized unigram estimation ($\sigma=0.1$), and set $T=2$ for \emph{Curated Video} data and $T=10$ for \emph{AMI Meeting} data. Table~\ref{tab:privacy} shows the WERs comparison results, where we can see that randomized FMP can offer reasonable utterance-level privacy protection with a relatively small cost in speech recognition accuracy. Particularly, the WER degradation is less than 1\% with $\epsilon\geq 0.5$.

\begin{table}[ht]
  \caption{WERs comparison among various randomized FMP methods ($\alpha=0.5$, $\beta=0.25$, $\sigma=0.1$) satisfying $(\epsilon, 0)$-DP.}
  \centering
  \resizebox{\columnwidth}{!}{%
  \begin{tabular}{l|l|l}
    \toprule
    & \multicolumn{2}{|c}{\bf{Evaluation Dataset}} \\
    \cmidrule(r){2-3}
    \emph{Method}& \emph{Curated Video} & \emph{AMI Meeting}  \\
    \midrule
    FMP w/ DP not satisfied & 7.67 & 31.02 \\
    \midrule
    FMP w/ $(2.0,0)$-DP satisfied\; & 7.67\;(+0.0\%) & 31.09\;(+0.2\%) \\
    FMP w/ $(1.0,0)$-DP satisfied\; & 7.67\;(+0.0\%) & 31.17\;(+0.5\%) \\    
    FMP w/ $(0.5,0)$-DP satisfied\; & 7.68\;(+0.1\%) & 31.26\;(+0.8\%) \\    
    FMP w/ $(0.1,0)$-DP satisfied\; & 7.74\;(+0.9\%) & 31.59\;(+1.8\%) \\    
    \bottomrule
  \end{tabular}
  }
  \label{tab:privacy}
\end{table}

\section{Conclusion}
\label{sec:summary}
In this work, we leverage FL to study fast marginal language model personalization with its application in ASR rescoring. Our approach could overcome the cold start and catastrophic forgetting issues confronting traditional federated fine-tuning, and efficiently learn personalized NNLMs on local devices. Experiments on two speech evaluation datasets show modest WER reductions. In the future, we plan to explore higher order of $n$-gram distribution for deeper personalization.
\section{Acknowledgements}
Thank Kshitiz Malik and Yutong Pang for discussions.

\vfill\pagebreak

\bibliographystyle{IEEEbib}
\bibliography{refs}

\begin{thebibliography}{10}

\bibitem{konevcny2016federated}
Jakub Kone{\v{c}}n{\`y}, H~Brendan McMahan, Daniel Ramage, and Peter
  Richt{\'a}rik,
\newblock ``Federated optimization: Distributed machine learning for on-device
  intelligence,''
\newblock {\em arXiv preprint arXiv:1610.02527}, 2016.

\bibitem{konevcny2016federated2}
Jakub Kone{\v{c}}n{\`y}, H~Brendan McMahan, Felix~X Yu, Peter Richt{\'a}rik,
  Ananda~Theertha Suresh, and Dave Bacon,
\newblock ``Federated learning: Strategies for improving communication
  efficiency,''
\newblock {\em arXiv preprint arXiv:1610.05492}, 2016.

\bibitem{mcmahan2017communication}
Brendan McMahan, Eider Moore, Daniel Ramage, Seth Hampson, and Blaise~Aguera
  y~Arcas,
\newblock ``Communication-efficient learning of deep networks from
  decentralized data,''
\newblock in {\em AISTATS}, 2017.

\bibitem{geyer2017differentially}
Robin~C Geyer, Tassilo Klein, and Moin Nabi,
\newblock ``Differentially private federated learning: A client level
  perspective,''
\newblock {\em arXiv preprint arXiv:1712.07557}, 2017.

\bibitem{chen2018federated}
Fei Chen, Mi~Luo, Zhenhua Dong, Zhenguo Li, and Xiuqiang He,
\newblock ``Federated meta-learning with fast convergence and efficient
  communication,''
\newblock {\em arXiv preprint arXiv:1802.07876}, 2018.

\bibitem{arnold2016suggesting}
Kenneth~C Arnold, Krzysztof~Z Gajos, and Adam~T Kalai,
\newblock ``On suggesting phrases vs. predicting words for mobile text
  composition,''
\newblock in {\em Proceedings of the 29th Annual Symposium on User Interface
  Software and Technology}, 2016.

\bibitem{leroy2019federated}
David Leroy, Alice Coucke, Thibaut Lavril, Thibault Gisselbrecht, and Joseph
  Dureau,
\newblock ``Federated learning for keyword spotting,''
\newblock in {\em Proc. ICASSP}, 2019.

\bibitem{kim2017federated}
Yejin Kim, Jimeng Sun, Hwanjo Yu, and Xiaoqian Jiang,
\newblock ``Federated tensor factorization for computational phenotyping,''
\newblock in {\em Proc. SIGKDD}, 2017.

\bibitem{xu2019federated}
Jie Xu and Fei Wang,
\newblock ``Federated learning for healthcare informatics,''
\newblock {\em arXiv preprint arXiv:1911.06270}, 2019.

\bibitem{goodman2001progress}
Joshua Goodman,
\newblock ``A bit of progress in language modeling,''
\newblock {\em Computer Speech and Language}, vol. 15, pp. 403--434, 2001.

\bibitem{mikolov2010recurrent}
Tom{\'a}{\v{s}} Mikolov, Martin Karafi{\'a}t, Luk{\'a}{\v{s}} Burget, Jan
  {\v{C}}ernock{\`y}, and Sanjeev Khudanpur,
\newblock ``Recurrent neural network based language model,''
\newblock in {\em Proc. Interspeech}, 2010.

\bibitem{chen2015improving}
Xie Chen, Xunying Liu, Mark~JF Gales, and Philip~C Woodland,
\newblock ``Improving the training and evaluation efficiency of recurrent
  neural network language models,''
\newblock in {\em Proc. ICASSP}, 2015.

\bibitem{xu2018pruned}
Hainan Xu, Tongfei Chen, Dongji Gao, Yiming Wang, Ke~Li, Nagendra Goel, Yishay
  Carmiel, Daniel Povey, and Sanjeev Khudanpur,
\newblock ``A pruned {RNNLM} lattice-rescoring algorithm for automatic speech
  recognition,''
\newblock in {\em Proc. ICASSP}, 2018.

\bibitem{brendan2018learning}
H.~Brendan McMahan, Daniel Ramage, Kunal Talwar, and Li~Zhang,
\newblock ``Learning differentially private recurrent language models,''
\newblock in {\em Proc. ICLR}, 2018.

\bibitem{popov2018distributed}
Vadim Popov, Mikhail Kudinov, Irina Piontkovskaya, Petr Vytovtov, and Alex
  Nevidomsky,
\newblock ``Distributed fine-tuning of language models on private data,''
\newblock in {\em Proc. ICLR}, 2018.

\bibitem{ji2019learning}
Shaoxiong Ji, Shirui Pan, Guodong Long, Xue Li, Jing Jiang, and Zi~Huang,
\newblock ``Learning private neural language modeling with attentive
  aggregation,''
\newblock in {\em IJCNN}. IEEE, 2019, pp. 1--8.

\bibitem{chen2019federated}
Mingqing Chen, Ananda~Theertha Suresh, Rajiv Mathews, Adeline Wong, Cyril
  Allauzen, Fran{\c{c}}oise Beaufays, and Michael Riley,
\newblock ``Federated learning of n-gram language models,''
\newblock {\em arXiv preprint arXiv:1910.03432}, 2019.

\bibitem{goodfellow2013empirical}
Ian~J Goodfellow, Mehdi Mirza, Da~Xiao, Aaron Courville, and Yoshua Bengio,
\newblock ``An empirical investigation of catastrophic forgetting in
  gradient-based neural networks,''
\newblock {\em arXiv preprint arXiv:1312.6211}, 2013.

\bibitem{dwork2006calibrating}
Cynthia Dwork, Frank McSherry, Kobbi Nissim, and Adam Smith,
\newblock ``Calibrating noise to sensitivity in private data analysis,''
\newblock in {\em Theory of Cryptography Conference}. Springer, 2006, pp.
  265--284.

\bibitem{dwork2014algorithmic}
Cynthia Dwork and Aaron Roth,
\newblock ``The algorithmic foundations of differential privacy,''
\newblock {\em Foundations and Trends in Theoretical Computer Science}, vol. 9,
  no. 3-4, pp. 211--407, 2014.

\bibitem{kneser1997language}
Reinhard Kneser, Jochen Peters, and Dietrich Klakow,
\newblock ``Language model adaptation using dynamic marginals,''
\newblock in {\em Fifth European Conference on Speech Communication and
  Technology}, 1997.

\bibitem{singh2017approximated}
Mittul Singh, Youssef Oualil, and Dietrich Klakow,
\newblock ``Approximated and domain-adapted {LSTM} language models for
  first-pass decoding in speech recognition,''
\newblock in {\em Proc. Interspeech}, 2017.

\bibitem{li2018recurrent}
Ke~Li, Hainan Xu, Yiming Wang, Daniel Povey, and Sanjeev Khudanpur,
\newblock ``Recurrent neural network language model adaptation for
  conversational speech recognition,''
\newblock in {\em Proc. Interspeech}, 2018.

\bibitem{shin2016generative}
Sungho Shin, Kyuyeon Hwang, and Wonyong Sung,
\newblock ``Generative knowledge transfer for neural language models,''
\newblock {\em arXiv preprint arXiv:1608.04077}, 2016.

\bibitem{bui2019federated}
Duc Bui, Kshitiz Malik, Jack Goetz, Honglei Liu, Seungwhan Moon, Anuj Kumar,
  and Kang~G Shin,
\newblock ``Federated user representation learning,''
\newblock {\em arXiv preprint arXiv:1909.12535}, 2019.

\bibitem{carletta2005ami}
Jean Carletta, Simone Ashby, Sebastien Bourban, Mike Flynn, Mael Guillemot,
  Thomas Hain, Jaroslav Kadlec, Vasilis Karaiskos, Wessel Kraaij, Melissa
  Kronenthal, et~al.,
\newblock ``The ami meeting corpus: A pre-announcement,''
\newblock in {\em International workshop on machine learning for multimodal
  interaction}. Springer, 2005, pp. 28--39.

\bibitem{graves2006connectionist}
Alex Graves, Santiago Fern{\'a}ndez, Faustino Gomez, and J{\"u}rgen
  Schmidhuber,
\newblock ``Connectionist temporal classification: labelling unsegmented
  sequence data with recurrent neural networks,''
\newblock in {\em Proc. ICML}, 2006.

\bibitem{vaswani2017attention}
Ashish Vaswani, Noam Shazeer, Niki Parmar, Jakob Uszkoreit, Llion Jones,
  Aidan~N Gomez, {\L}ukasz Kaiser, and Illia Polosukhin,
\newblock ``Attention is all you need,''
\newblock in {\em Proc. NeurIPS}, 2017.

\bibitem{dwork2008differential}
Cynthia Dwork,
\newblock ``Differential privacy: A survey of results,''
\newblock in {\em International Conference on Theory and Applications of Models
  of Computation}. Springer, 2008, pp. 1--19.

\end{thebibliography}

\end{document}